\documentclass[twoside,twocolumn]{article}

\usepackage{fitee}
\usepackage{fancyvrb}

\usepackage[hidelinks, breaklinks = true]{hyperref}  

\addto{\captionsenglish}{%
  \renewcommand{\figurename}{Fig.}
}

\long\def\c#1{{\footnotesize{\fontfamily{pcr}\selectfont\textbf{#1}}}}

\begin{document}

\title{Task Planning in Robotics: an Empirical Comparison of PDDL-based and ASP-based Systems}

\author[1]{Yuqian Jiang}%
\author[$\dagger$2]{Shiqi Zhang}%
\author[3]{Piyush Khandelwal}%
\author[1]{Peter Stone}%

\affil[1]{Department of Computer Science, The University of Texas at Austin}
\affil[2]{Department of Computer Science, SUNY Binghamton}
\affil[3]{Cogitai, Inc.}

\shortauthor{Jiang et al.}	

\authmark{}



\corremailA{jiangyuqian@utexas.edu} 
\corremailB{szhang@cs.binghamton.edu}
\corremailC{piyushk@gmail.com}
\corremailD{pstone@cs.utexas.edu}
\emailmark{}	

\dateinfo{Received mm.\ dd, yyyy;	Revision accepted mm.\ dd, yyyy;    Crosschecked mm.\ dd, yyyy}

\abstract{Robots need task planning algorithms to sequence actions toward accomplishing goals that are impossible through individual actions. 
Off-the-shelf task planners can be used by intelligent robotics practitioners to solve a variety of planning problems. 
However, many different planners exist, each with different strengths and weaknesses, and there are no general rules for which planner would be best to apply to a given problem. 
In this article, we empirically compare the performance of state-of-the-art planners that use either the Planning Domain Description Language (\textsc{PDDL}), or Answer Set Programming (\textsc{ASP}) as the underlying action language. 
\textsc{PDDL} is designed for task planning, and \textsc{PDDL}-based planners are widely used for a variety of planning problems. 
\textsc{ASP} is designed for knowledge-intensive reasoning, but can also be used for solving task planning problems. 
Given domain encodings that are as similar as possible, we find that \textsc{PDDL}-based planners perform better on problems with longer solutions, and \textsc{ASP}-based planners are better on tasks with a large number of objects or in which complex reasoning is required to reason about action preconditions and effects. 
The resulting analysis can inform selection among general purpose planning systems for particular robot task planning domains. 
}


\doi{10.1631/FITEE.1000000}	
\code{A}
\clc{TP}


\publishyear{2019}
\vol{??}
\issue{?}
\pagestart{0}
\pageend{0}




\articleType{}

\maketitle

\section{Introduction}
\label{sec:intro}

In a general purpose planning system, task planning problems are tackled by problem-independent solvers based on a description of the domain in a declarative language.
Such planning systems are extremely useful in application domains where many different planning goals need to be accomplished, or the domain description evolves over time. 
For instance, in an application domain such as robotics, a mobile service robot may need to solve planning tasks such as collecting documents, making deliveries, or providing navigation assistance to visitors~\citep{cambon2009hybrid,erdem2012answer,khandelwal2017bwibots}.
It is convenient to achieve all these tasks using knowledge declared in a single description of the domain, and general purpose planning systems are well suited to the task.

In order to design a general purpose planning system, a declarative language for formalizing the domain first needs to be selected, followed by the selection of a suitable solver which supports this language. 
Many different factors affect this selection process.
Every language has its limitations in representing task planning problems, and given a particular language, specific language-dependent techniques may need to be employed to succinctly formalize a particular planning problem. 
For instance, not all languages support default reasoning, which might bring difficulties in formalizing some planning problems. 
A solver is also typically tied to a particular language, but may not support all features in that language, requiring careful construction of the domain description using only supported features.
Additionally, the properties of the domain can affect how quickly a given pair of language and solver can solve planning problems. 
For instance, some domains include many objects and their properties, which can be challenging to some planning systems. 
Finally, given a language, a solver and a planning problem, there can be many ways of formalizing the problem using the language.
For these reasons, careful consideration needs to be given to the selection of language and solver.

This article aims to help in the language selection process, given a task planning problem at hand.
Specifically, we compare two declarative languages: the Planning Domain
Definition Language (\textsc{PDDL})~\citep{mcdermott1998pddl}, the most
popular language in the planning community, and Answer Set Programming (\textsc{ASP})~\citep{gelfond2014knowledge,lifschitz2008answer}, a popular
general knowledge representation and reasoning (KRR) language that has been
recently used in a variety of task planning
problems~\citep{lifschitz2002answer,yang2014planning,erdem2016applications}, including robotics~\citep{erdem2018applications}. 
\textsc{PDDL} was created for the explicit purpose of solving planning problems, whereas
the development of \textsc{ASP} has focused on a broader set of reasoning tasks, such as
inference and diagnosis, as well as planning.

The main contribution of this article is, within the context of robotics, a comparison of planning time between \textsc{ASP}-based and \textsc{PDDL}-based task planners when both are used to model the same domain.
Evaluation is performed across three different benchmark problems.
While it may be possible to construct ASP and PDDL planners specifically suited to these specific benchmarks, domain-independent solvers are compared in this article.
Although planner performance can be sensitive to domain encoding, we take care, to the extent possible, to encode the domains similarly in each language.
The benchmark problems consist of the Blocks World and Hiking problems from the International Planning Competition (\textsc{IPC})~\citep{coles2012survey}, as well as a variant of the Robot Navigation problem~\citep{yang2014planning,zhang2015mobile}.
The Robot Navigation problem typically requires more complex reasoning about action preconditions and effects than problems in the \textsc{IPC}, which focus on generating long plans efficiently.
Various properties of the domain or task in these benchmarks are also varied during evaluation to analyze the effect on planning time.
The goal of this article is to help a robot planning practitioner understand the effects of specific domain properties to aid the choice of language selection for general purpose planning.

We hypothesize that current state-of-the-art \textsc{PDDL}-based planners perform better on tasks with long solutions, and \textsc{ASP}-based planners tend to perform better on shorter tasks with a large number of objects. 
The hypothesis is confirmed in all three benchmark domains. 
We also hypothesize that \textsc{ASP}-based planners outperform \textsc{PDDL}-based planners in domains where \emph{complex reasoning} (specified in Section~\ref{sec:navigation}) is required to reason about action preconditions and effects. 
This is observed in a controlled experiment of the Robot Navigation and Blocks World domains.
To the best of our knowledge, this is the first work on empirical comparisons between \textsc{PDDL}-based and \textsc{ASP}-based planning systems, and can serve as a useful reference to robot planning practitioners.

\section{Background}
\label{sec:background}

Research in task planning dates back to one of the earliest research areas in Artificial Intelligence. 
Since the development of STRIPS~\citep{fikes1971strips} (as part of the \emph{Shakey} robot project), many languages have been developed for representing task planning domains. 
Such languages typically need to describe actions' preconditions and effects, and are commonly known as \emph{action languages}. 
A summary of some early action languages is available~\citep{gelfond1998action}. 

In order to plan for real-world problems (such as robot systems), action languages first need to be capable of formally representing complex planning domains. 
Some of the recent research in task planning is focused on developing languages that improve the representation capability of action languages~\citep{giu04,lee13,babb2015action}. 
Planning using these action languages usually requires a translation to more general knowledge representation and reasoning (KRR) languages such as \textsc{ASP}. 
A planning paradigm for \textsc{ASP} was proposed~\citep{lifschitz2002answer}, and has been used in many real-world applications~\citep{chen2010developing,erdem2012answer,yang2014planning}.
In this article, we follow the same planning paradigm when encoding domains in \textsc{ASP}.

In parallel, another line of research in task planning aims at more efficient planning algorithms and their implementations.
\textsc{PDDL}~\citep{mcdermott1998pddl} was developed as a common formalism with the goal of allowing more direct comparison of planning algorithms and implementations.  
Since then, many efficient search algorithms have been developed for task planning problems, such as Fast-Forward~\citep{hoffmann2001ff} and Fast-Downward~\citep{helmert2006fast}.
These algorithms have publicly available implementations including SAYPHI~\citep{de2007using}, LAMA~\citep{richter2011lama} and FDSS~\citep{helmert2011fast}.


\textsc{PDDL} requires \emph{axioms}, in the form of logical formulas, for reasoning
within a situation (whereas action descriptions are used for reasoning across
successive situations). A fundamental difference between \textsc{PDDL} and \textsc{ASP} is on
their (non)monotonicity property. The axiom-based reasoning in \textsc{PDDL} is monotonic
in the sense of logical reasoning, meaning that previously achieved
conclusions remain when new information becomes available. In contrast, \textsc{ASP} is
nonmonotonic, so it allows removal of previously achieved conclusions given new
information.
The nonmonotonic property of \textsc{ASP} makes it useful in tasks that require default reasoning and reasoning about inertial facts. 
Existing research has studied translating \textsc{PDDL} programs to \textsc{ASP}~\citep{gebser2011plasp}, and applying axioms extracted from PDDL programs to ASP-based planning~\citep{ICAPS1715730}. 
In particular, robotics researchers have developed robot navigation algorithms that switch between ASP-based and PDDL-based planning systems on mobile service robots~\citep{lo2018petlon}. 
However, none of this research conducted empirical comparisons over the performances of the state-of-the-art \textsc{PDDL}-based and \textsc{ASP}-based planning systems. 

Action languages can be further categorized as \emph{Action Description Languages} and \emph{Action Query Languages}~\citep{lifschitz1997two}. 
Action description languages focus on specifying the transition system. 
Given a transition system, action query languages are used for reasoning about properties of trajectories, such as to reason about history, non-determinism, or both for diagnosis purposes. 
From the perspective of design purposes, PDDL is an action description language, and ASP is an action query language, though their implementations oftentimes support both description and query functionalities. 
At the same time, action language systems are usually implemented by compilation. 
For instance, \emph{Coala}~\citep{gebser2010coala} is one such compilation system that provides compilation techniques for several action languages. 

There is existing research on predicting planning time using features of domains and problems~\citep{fawcett2014improved}, or more generally on predicting time required to solve a problem~\citep{leyton2002learning}. 
These methods can be used to help a planning practitioner to estimate the difficulty of a planning problem, after planning language and system have been selected. 
In contrast, this work aims at analyzing what domain properties affect the performances of existing planning systems, and can serve as a reference on the selection of action languages used for encoding planning problems.

\section{Domain Formalization}
\label{sec:formal}

In this section, we introduce the three benchmark domains, namely Robot Navigation, Blocks World and Hiking, and formally describe each in both \textsc{ASP} and \textsc{PDDL}.

It may be possible that different styles of encoding would result in different planning times. To ensure the conclusion is general and fair, we select the benchmark domains from a variety of origins, and follow the encoding in existing literatures. Blocks World and Hiking have been used in the \textsc{IPC}, and \textsc{PDDL}-based planners have been specifically designed to solve problems in the \textsc{IPC}.
The \textsc{PDDL} versions are used as is, and translated into equivalent \textsc{ASP} code.
Robot Navigation is a domain used to demonstrate planning using an ASP description~\citep{yang2014planning}, and contains properties such as recursive action effects that are missing from IPC domains.
We translate the complex reasoning rules of the Robot Navigation domain from \textsc{ASP} as \textsc{PDDL} axioms.

The languages themselves require certain differences in domain encodings, and direct comparisons are complicated by the fact that typically the people doing the encoding have greater familiarity with one language or the other. 
We acknowledge that different encodings may be more suitable to each planner, but it is infeasible to control the encoding style in practice. 
In this work, \emph{we ensure the fairest comparison possible by enforcing in all translations that both the \textsc{ASP} and \textsc{PDDL} versions for a given domain have exactly the same set of predicates and actions, along with the same preconditions and effects.}
Specifically, an action should be allowed to execute on the same set of states, and it should make the same change to the state, regardless of the language.
Consequently, as we use optimal \textsc{ASP} and \textsc{PDDL} planners in the experiments, they generate identical plans, and only planning times need to be compared.
A detailed explanation of how \textsc{ASP} can be used for planning is available in previous work~\citep{lifschitz2002answer}.


\subsection{Robot Navigation}
\label{sec:navigation}

The robot navigation domain differs from classical planning domains in the \textsc{IPC} in that complex reasoning needs to be performed to ensure that action preconditions are met, and that action effects are executed correctly.
Specifically, this domain features action effects that require recursive reasoning to compute the final state of the world.
In this domain, a mobile robot navigates an office floor which consists of a set of rooms that are connected to one another.
Rooms can be connected to one another via doors, and closed doors need to be opened by the robot before it can pass through. 
Alternatively, rooms can be directly connected to one another such that access is always possible from any location in one room to any location in the other.

The robot has the following perception and actuation modules available.
Using a low-level controller, the robot can traverse to any room from its current location if its path is not blocked by a closed door, and this navigation can be encoded by a single high-level symbolic action \c{goto}.
Furthermore, the robot has some means of opening a closed door when it is next to it, either by enlisting human aid, or using a robot arm to open the door.
On the perception side, the robot can sense its location, whether or not it is next to a door, and whether or not a door is open.

The domain knowledge can be formalized in \textsc{ASP} by statements defined using the following predicates:

\begin{itemize}

  \item \c{hasdoor(R,D)}: This predicate specifies that room \c{R} has door \c{D} to move to an adjacent location.
    Statements expressed using \c{hasdoor} are specified during initialization, and do not change over time.
 The \textsc{PDDL} expression is \c{(hasdoor ?r - room ?d - door)}.

\item \c{connected(R1,R2)}: The \c{connected} predicate indicates that room \c{R1} is directly connected to room \c{R2} without a door.
Similar to \c{hasdoor}, this predicate is used during initialization to describe directly connected locations.
In \textsc{PDDL}, statements are described as \c{(connected ?r1 - room ?r2 - room)}.

\item \c{acc(R1,R2)}: \c{acc} specifies that room \c{R1} is accessible from room \c{R2} via a single navigation action executed by a low level controller.
Intuitively, any two rooms that are not separated by a closed door are accessible, i.e.~the low level controller can navigate from one room to another.

\c{R1} is accessible from \c{R2} if \c{R1} is directly connected to \c{R2}, or \c{R1} and \c{R2} share the same door \c{D} which is open.
  Furthermore, \c{acc} is both commutative, i.e.~\c{R1} is accessible from room \c{R2} if \c{R2} is accessible from \c{R1}, as well as associative, i.e.~if both \c{R1} and \c{R2} are accessible from \c{r3}, then they're accessible from one another.
This associative property requires a recursive definition:

\begin{quote}
\begin{scriptsize}
\begin{Verbatim}[fontfamily=courier, fontseries=b]
acc(R1,R2,n) :- connected(R1,R2).
acc(R1,R2,n) :- open(D,n), 
                hasdoor(R1,D), 
                hasdoor(R2,D).
acc(R1,R2,n) :- acc(R2,R1,n).
acc(R1,R2,n) :- acc(R1,R3,n), 
                acc(R3,R2,n).
\end{Verbatim}
\end{scriptsize}
\end{quote}

The recursive formulation of \c{acc} can be expressed in \textsc{PDDL} using derived predicates:
\begin{quote}
\begin{scriptsize}
\begin{Verbatim}[fontfamily=courier, fontseries=b]
(:derived (acc ?r1 - room ?r2 - room)
  (or (connected ?r1 ?r2) 
      (exists (?d - door) (and (open ?d) 
                               (hasdoor ?r1 ?d)
                               (hasdoor ?r2 ?d)))
      (acc ?r2 ?r1)
      (exists (?r3 - room) (and (acc ?r1 ?r3)
                                (acc ?r3 ?r2)))))
\end{Verbatim}
\end{scriptsize}
\end{quote}

\item \c{at(R,n)}: \c{at} is used to specify that the robot is at room \c{R} at timestep \c{n} (of the high-level plan).
  This predicate is inertial, i.e.~the robot remains in room \c{R} if there is no evidence showing it is not in room \c{R} anymore, and this property is specified as:
\begin{quote}
\begin{scriptsize}
\begin{Verbatim}[fontfamily=courier, fontseries=b]
at(R,n) :- at(R,n-1), not -at(R,n).
\end{Verbatim}
\end{scriptsize}
\end{quote}

In \textsc{PDDL}, the predicate is expressed as \c{(at ?r - room)}, and all predicates are inertial by default.

\item \c{open(D,n)}: Door \c{D} is open at step \c{n}. 
  \c{door} is inertial, i.e.~the robot believes that a door will stay in the same state unless sensed differently.
  In \textsc{ASP}, the inertial property for this predicate is represented as:
\begin{quote}
\begin{scriptsize}
\begin{Verbatim}[fontfamily=courier, fontseries=b]
open(D,n) :- open(D,n-1), not -open(D,n).
\end{Verbatim}
\end{scriptsize}
\end{quote}

In \textsc{PDDL}, \c{open} is expressed as \c{(open ?d - door)}.

\item \c{canopen(D,n)}: The robot can open door \c{D} if it is right next to it.
  \c{canopen} is the action effect of approaching a door, and a precondition before the door can be opened.
  \c{canopen} is not inertial.
  In \textsc{PDDL}, the predicate is expressed as \c{(canopen ?d - door)}.
  
\item \c{visited(R,n)}: Once the robot visits a room, the visited fluent for that room remains true until the end of planning process.
  \c{visited} is used to describe goal conditions. The persistence property is expressed in \textsc{ASP} as:
\begin{quote}
\begin{scriptsize}
\begin{Verbatim}[fontfamily=courier, fontseries=b]
visited(R,n) :- visited(R,n-1).
\end{Verbatim}
\end{scriptsize}
\end{quote}

\c{(visited ?r - room)} expresses \c{visited} in \textsc{PDDL}.

\end{itemize}

There are three actions in the domain: \c{goto}, \c{approach}, and \c{opendoor}, with the following definitions:

\begin{itemize}

  \item \c{goto(R2,n)}: This action specifies that the robot should navigate to room \c{R2} at timestep \c{n} in the high-level plan.
    The precondition for this action is that robot must be in a room \c{R1} from which \c{R2} is accessible.
    Once the robot goes to room \c{R2}, the goal condition \c{visited} is set to true for that room as well.
    The \textsc{ASP} rules defining the action preconditions and effects are as follows:
\begin{quote}
\begin{scriptsize}
\begin{Verbatim}[fontfamily=courier, fontseries=b]
:- goto(R2,n), at(R1,n-1), not acc(R1,R2,n-1).
at(R2,n)      :- goto(R2,n).
-at(R1,n)     :- goto(R2,n), at(R1,n-1), R1 != R2.
visited(R2,n) :- goto(R2,n).
\end{Verbatim}
\end{scriptsize}
\end{quote}

The same description in \textsc{PDDL} is expressed as follows:
\begin{quote}
\begin{scriptsize}
\begin{Verbatim}[fontfamily=courier, fontseries=b]
(:action goto
  :parameters (?r2 - room)
  :precondition (exists (?r1 - room) 
                        (and (at ?r1) 
                        (acc ?r1 ?r2)))
  :effect (and (at ?r2)
               (forall (?r1 - room) 
                       (when (at ?r1) 
                             (not (at ?r1))))
               (visited ?r2)
               (forall (?d1 - door) 
                       (not (canopen ?d1)))))
\end{Verbatim}
\end{scriptsize}
\end{quote}
The action effects in \textsc{PDDL} have an additional statement than \textsc{ASP} to indicate that \c{canopen} is not inertial.

  \item \c{approach(D,n)}: This action specifies that the robot should approach door \c{D}.
    The action is only executable when the robot is in a room that has door \c{D}.
    After executing this action, the robot can open door \c{D}.
In \textsc{ASP}, this action is expressed as:

\begin{quote}
\begin{scriptsize}
\begin{Verbatim}[fontfamily=courier, fontseries=b]
:- approach(D,n), at(R1,n-1), not hasdoor(R1,D).
canopen(D,n) :- approach(D,n).
\end{Verbatim}
\end{scriptsize}
\end{quote}

In \textsc{PDDL}, this action is expressed as:
\begin{quote}
\begin{scriptsize}
\begin{Verbatim}[fontfamily=courier, fontseries=b]
(:action approach
  :parameters (?d - door)
  :precondition (exists (?r1 - room) 
                        (and (at ?r1) 
                             (hasdoor ?r1 ?d)))
  :effect (and (canopen ?d)
               (forall (?d1 - door) 
                       (when (not (= ?d1 ?d)) 
                             (not (canopen ?d1))))))
\end{Verbatim}
\end{scriptsize}
\end{quote}

The action effects in \textsc{PDDL} both express that door \c{d} can be opened, and no other doors in the domain can be opened without approaching them first.

\item \c{opendoor(D,n)}: This action allows the robot to open door \c{D} if \c{canopen(D)} is true.
Opening an open door does not change the state of the world. 
This action is represented in \textsc{ASP} as:
\begin{quote}
\begin{scriptsize}
\begin{Verbatim}[fontfamily=courier, fontseries=b]
:- opendoor(D,n), not canopen(D,n-1).
open(D,n) :- opendoor(D,n).
\end{Verbatim}
\end{scriptsize}
\end{quote}

In \textsc{PDDL}, this action is represented as:
\begin{quote}
\begin{scriptsize}
\begin{Verbatim}[fontfamily=courier, fontseries=b]
(:action opendoor
  :parameters (?d - door)
  :precondition (canopen ?d)
  :effect (and (open ?d)
               (forall (?d1 - door) 
                       (not (canopen ?d1)))))
\end{Verbatim}
\end{scriptsize}
\end{quote}

\end{itemize}

In all three actions, a specific action effect in \textsc{PDDL} describes the non-inertial property of \c{canopen}.
This effect is not required in \textsc{ASP} because the \c{canopen} is not inertial.

The goal in the Robot Navigation domain is to visit a randomly selected set of rooms.
In order to visit a room, the robot needs to recursively reason about which rooms are accessible from one another. 
If a door in its path is closed, the robot needs to explicitly approach the door and execute an \c{opendoor} action.
Whenever \c{opendoor} is executed, the direct accessibility of rooms changes, and this action effect needs to be computed recursively.
This recursive property of the domain differentiates it from traditional \textsc{IPC} planning domains such as Blocks World and Hiking.

\subsection{Blocks World}
In the Blocks World domain, the goal is to move a set of stackable blocks from one configuration to another using a robot hand.
We use the official domain definition in \textsc{IPC-2011} as the \textsc{PDDL} domain description, and translate the domain to \textsc{ASP}.
The full description of the \textsc{IPC} domain is available online
in \textsc{PDDL}\footnote{\url{http://pastebin.com/raw/b07aMTJB}} and \textsc{ASP}\footnote{\url{http://pastebin.com/raw/SAqM3xbF}}.
We only describe the translation of the \c{pick-up} action to \textsc{ASP} as an illustrative example; the other actions (\c{put-down}, \c{stack}, and \c{unstack}) follow similarly.
%
%
\c{pick-up} allows a robot to pickup a block that has no blocks underneath it (designated by \c{ontable}).
Furthermore, it should also have no blocks stacked on top of it (designated by \c{clear}).
Finally, a block can only be picked up if the robot hand is empty (designated by \c{handempty}).
The effect of the action is that the robot hand is holding the block, and all preconditions for the action become false.
This action description translated to \textsc{ASP} looks as follows:

\begin{quote}
\begin{scriptsize}
\begin{Verbatim}[fontfamily=courier, fontseries=b]
:- pickup(B,n), not clear(B,n-1).
:- pickup(B,n), not ontable(B,n-1).
:- pickup(B,n), not handempty(n-1).
-ontable(B,n) :- pickup(B,n).
-clear(B,n)   :- pickup(B,n).
-handempty(n) :- pickup(B,n).
holding(B,n)  :- pickup(B,n).
\end{Verbatim}
\end{scriptsize}
\end{quote}

The first three statements specify the same action preconditions specified in the \textsc{PDDL} description, and the last four statements specify the same action effects as specified in the \textsc{PDDL} description.

The extended version of the Blocks World domain introduces a recursively defined predicate: \c{above}~\citep{THIEBAUX200538}. The \textsc{PDDL} definition is as follows:

\begin{quote}
\begin{scriptsize}
\begin{Verbatim}[fontfamily=courier, fontseries=b]
(:derived (above ?x ?y)
    (or (on ?x ?y)
        (exists (?z) (and (on ?x ?z) (above ?z ?y)))))
\end{Verbatim}
\end{scriptsize}
\end{quote}

The predicate is defined in \textsc{ASP} as:
\begin{quote}
\begin{scriptsize}
\begin{Verbatim}[fontfamily=courier, fontseries=b]
above(X,Y,n) :- on(X,Y,n).
above(X,Y,n) :- on(X,Z,n), above(Z,Y,n).
\end{Verbatim}
\end{scriptsize}
\end{quote}

The complete domain description in \textsc{PDDL}\footnote{\url{http://pastebin.com/raw/FwZgGmZf}} and \textsc{ASP}\footnote{\url{http://pastebin.com/raw/9D2PuNze}} are online. We use both the original and the extended versions of the Blocks World domain in experiments.

\subsection{Hiking}
We select the hiking domain, new in \textsc{IPC-2014}, as our third benchmark domain.
The hiking domain features negative preconditions.
We use the official \textsc{PDDL} formalization in \textsc{IPC}, and an equivalent definition in \textsc{ASP} for this study.

In short, the purpose of this domain is to arrange activities for a number of couples so each couple can hike to their destination with a tent ready.
A hiking problem specifies connections between places, and initial locations of couples, cars, and tents.
A typical plan transports and sets up tents at the destination, drives each couple to the starting point of their hike, and then has them walk together along the hike.
The complete description of the Hiking domain is available online for both \textsc{PDDL}\footnote{\url{http://pastebin.com/raw/v3wkv57W}} and \textsc{ASP}\footnote{\url{http://pastebin.com/raw/Dw1BwG0Z}}.


%

\section{Experiments}
\label{sec:experiments}

The experiments in this section are designed to compare planning times when the domains and problems formalized in the previous section are optimally solved using state-of-the-art solvers. Our encoding strategy and the award-winning optimal planning systems ensure that all planners generate the same plan given the same pair of domain and problem. 

We select FastDownward~\citep{helmert2006fast} with the setting \textsc{FDSS-1}~\citep{helmert2011fast}, which had the highest score in the sequential optimization track at \textsc{IPC} 2011. 
It should be noted that the versions of IPC after 2011 do not take into account or announce the time needed to solve each problem. 
Instead, a hard time constraint (e.g., 30 minutes) is given to all planners. 
As a result, most \textsc{PDDL} planners always use the maximum time allowed to avoid reporting suboptimal or incorrect solutions (which is greatly penalized), and are therefore unfit for a comparison of planning times. 
The Robot Navigation domain and the extended Blocks World domain require derived predicates. Since none of the planners in the optimization track support derived predicates, we use FastDownward with the setting \textsc{LAMA}-2011~\citep{richter2011lama}, the winner of sequential satisficing track in 2011.
We use version 4.5.4 of Clingo~\citep{GebserKKS14} in incremental mode as the \textsc{ASP} solver. Clingo is an Answer Set solving system that integrates Clasp, the winner of the fifth Answer Set Programming Competition in 2015~\citep{calimeri2016design}.

Planner performances are evaluated on a general-purpose High Throughput Computing (HTC) cluster that is operated by the Department of Computer Science at the University of Texas at Austin. 
We filter out machines with less than 8GB memory in the experiments, resulting in more than ten machines with different hardware configurations (e.g., memory ranging from 8GB to more than 500GB). 
All data points are averaged across at least 10 trials to reduce noise, and standard deviations are reported. Given the wide range of hardware configurations and statistical analysis, we aim to conclude with observations that are generally valid and useful to practitioners.

\begin{figure*}[t]
    \begin{center}\hspace{0em}
    \includegraphics[width=.7\textwidth]{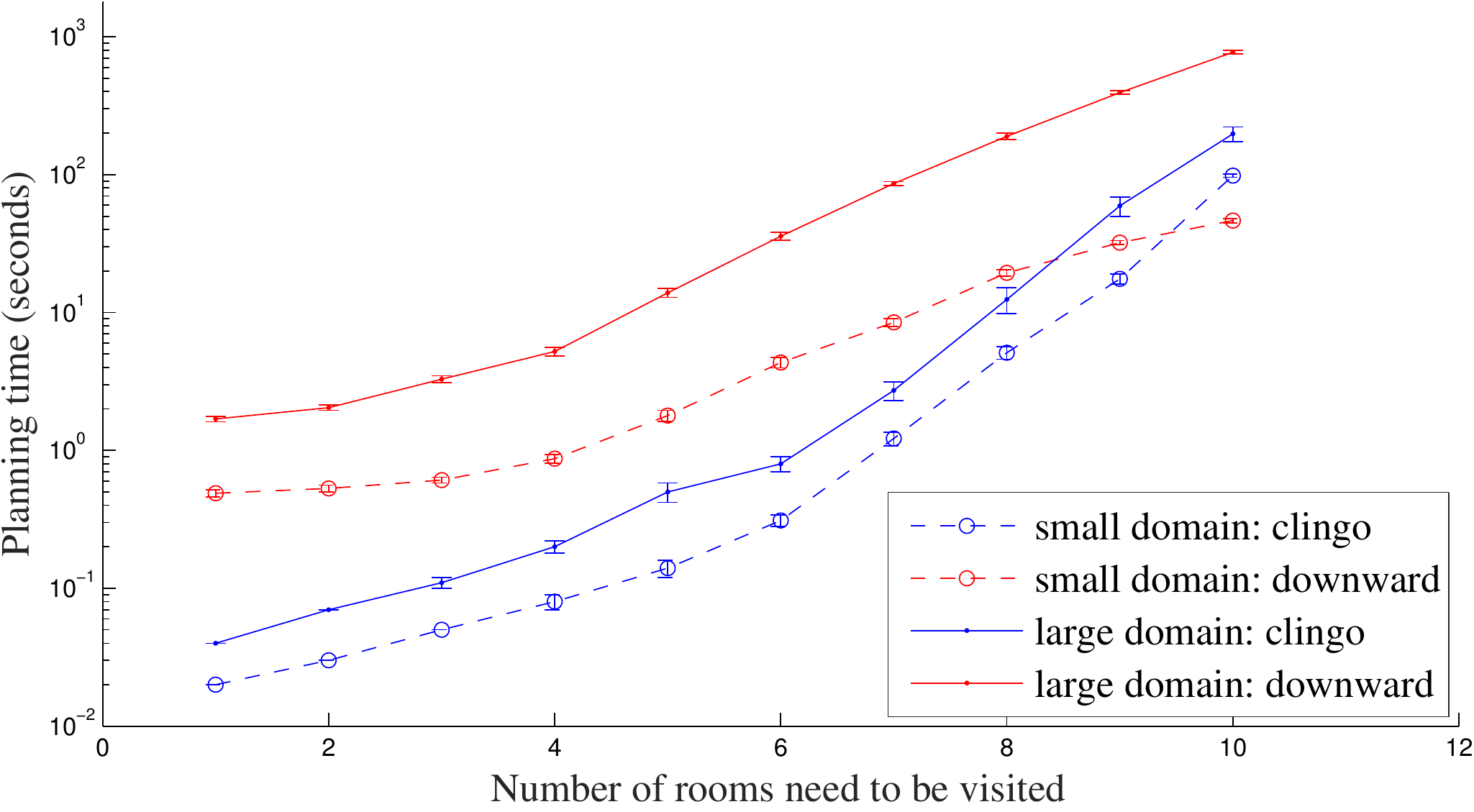} 
    \caption{Robot Navigation: Small Domain - 10 rooms, Large Domain - 15 rooms.
40\% of rooms are connected via doors, and the rest are directly accessible from the corridor.}
    \label{fig:navigation}
    \end{center}
    \vspace{-10pt}
\end{figure*}

During evaluations, various domain characteristics are also changed to measure the difference in performance of different planning paradigms.
It should be noted that we are generally more interested in comparing \emph{sensitivities} of planning systems (instead of comparing individual data points) given different domain characteristics. 
The sensitivity that can be reflected by the ``trend'' of a series of data points is typically more robust to implementation details of planning systems such as programming languages and compilers. 
The goal of these evaluations is to test the hypothesis that \textsc{PDDL}-based approaches work better in situations where the generated plans have many steps, and \textsc{ASP}-based approaches work better in situations where the domain is large or substantial reasoning is required at every step of the plan to compute the world state.



\paragraph{Robot Navigation} Results from the Robot Navigation domain are shown in Figure~\ref{fig:navigation}.
Two different versions of the domain have been created.
The small domain has 10 rooms, and each of them is connected a central corridor.
4 rooms are connected via doors, and the rest are directly connected. 
In contrast, the large domain has 15 rooms where 6 rooms are connected via doors to the corridor.

The robot is initially located in the corridor, and as its goal, needs to visit a number of rooms (represented on the x-axis in Figure~\ref{fig:navigation}) that are randomly selected in each trial. Since rooms that are connected by doors take more steps to visit, we increase the number of trials (to 50), on which we compute the average planning time for each data point. 
Regardless, the number of rooms in the goal is positively correlated with the plan length.
The planning time for each planner is represented on the y-axis (log scale, same for all following figures).

We can observe that the red curves have smaller slopes, since they intersect (or will do so) with the blue curves.
This confirms our hypothesis that \textsc{PDDL} planners are better at solving planning problems which require a large number of steps.
We can also observe that the gap between red curves is larger than the gap between blue curves, even on logarithmic scale.
This observation again supports the hypothesis that \textsc{ASP}-based planning is less sensitive to object scaling.

Furthermore, \textsc{ASP}-based planning is much faster than \textsc{PDDL}-based planning when the number of rooms is less than eight, and finishes within a reasonable amount of time. 
This is especially useful in domains such as robotics where fast real-time operation is necessary.
This better performance is probably a consequence of recursive action effects embedded in the domain. We further verify this observation in a controlled experiment using the Blocks World domain.

\begin{figure*}[t]
    \begin{center}\hspace{0em}
    \includegraphics[width=.7\textwidth]{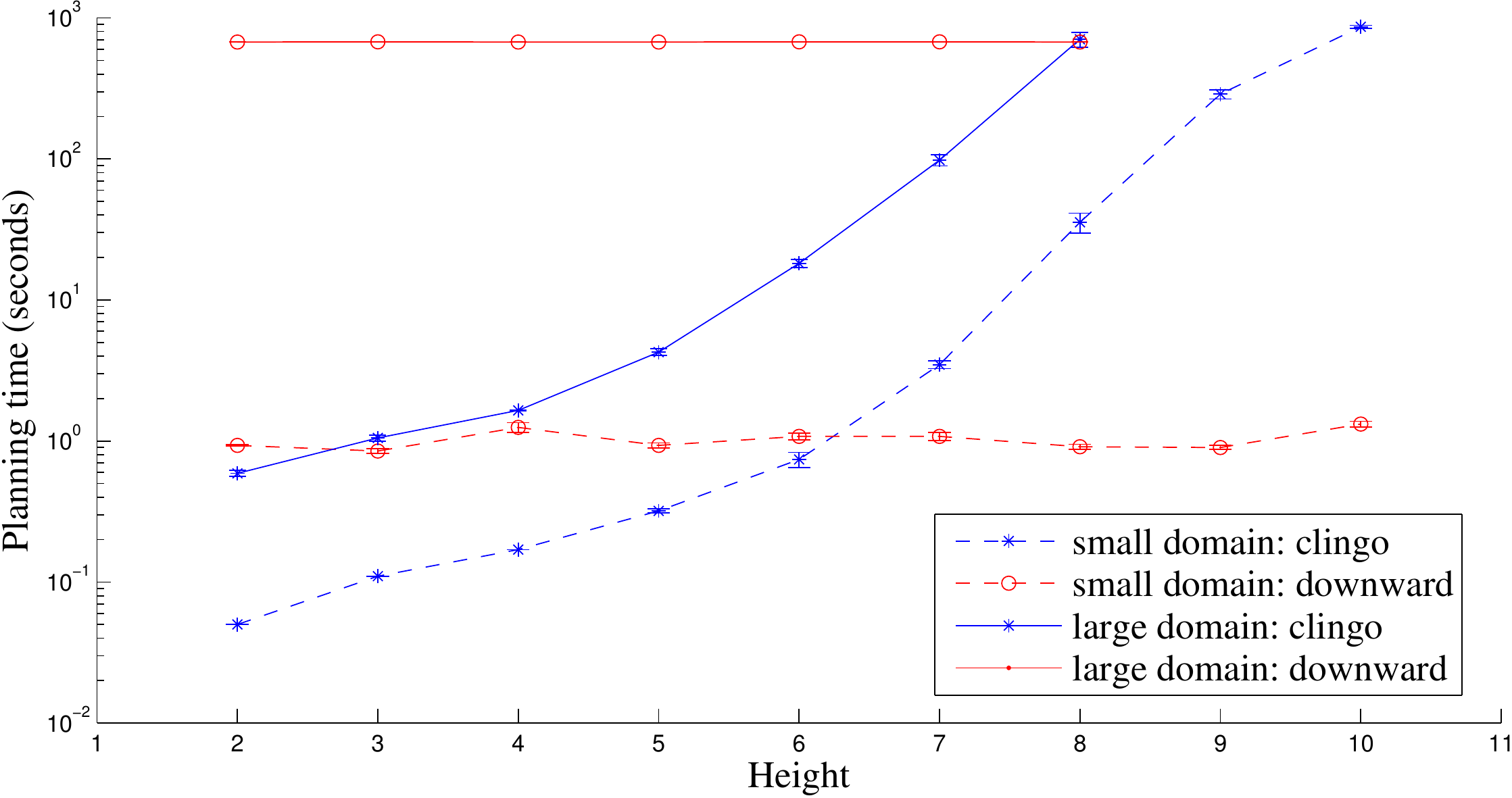} 
  \caption{Blocks World: Small Domain - 15 blocks, Large Domain - 60 blocks. We use a timeout of 1800 seconds (same as IPC). The graph only plots configurations where all trials of both planners finished before timeout (same for all following experiments). }    
    \label{fig:blocksworld}
    \end{center}
    \vspace{-10pt}
\end{figure*}

\begin{figure*}[t]
    \begin{center}\hspace{0em}
    \includegraphics[width=.7\textwidth]{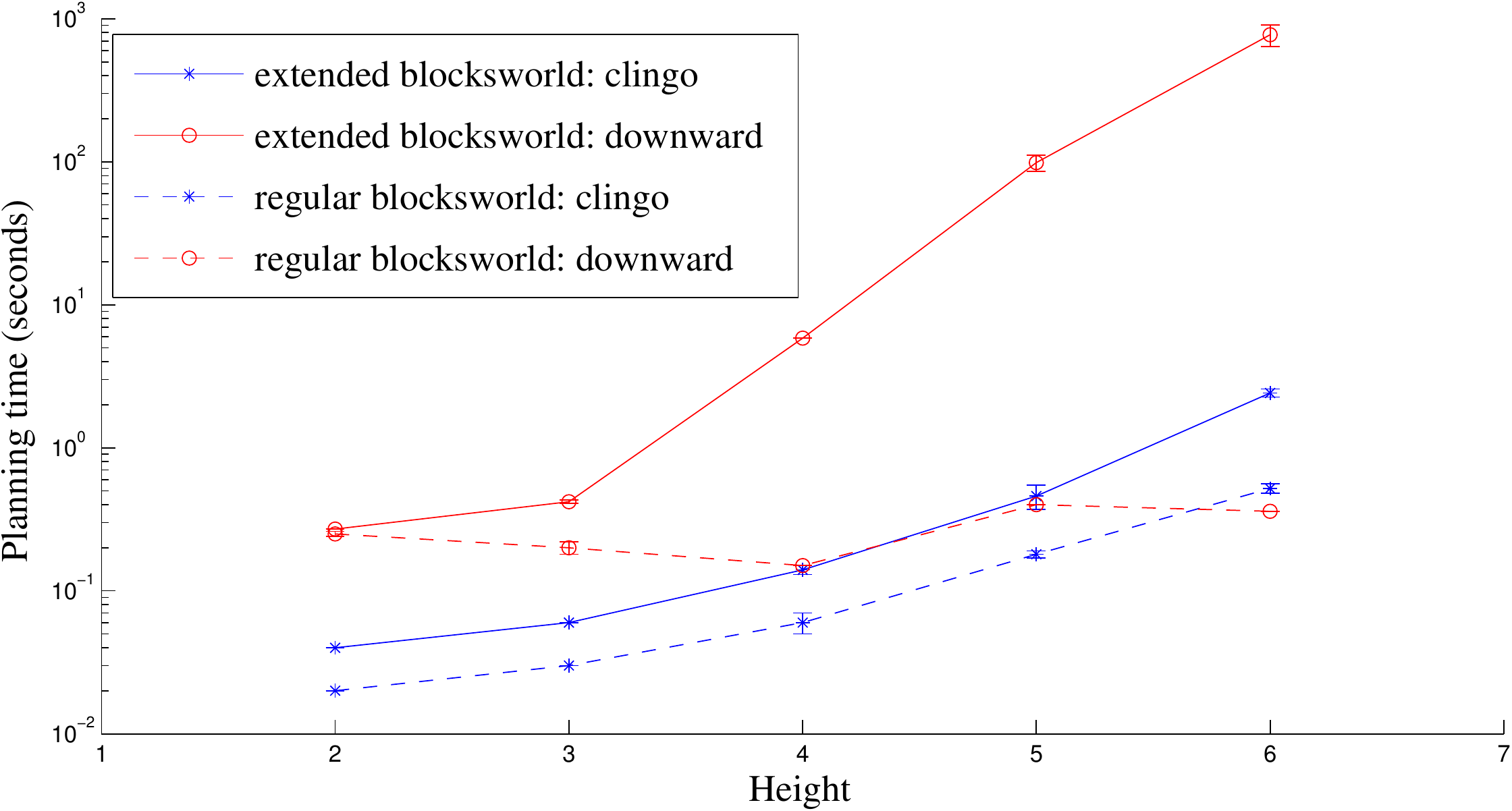} 
  \caption{Blocks World: Regular Domain vs. Extended Domain.}    
    \label{fig:blocks_extended}
    \end{center}
    \vspace{-10pt}
\end{figure*}


\begin{figure*}[t]
  \begin{center}
    \includegraphics[width=.95\textwidth]{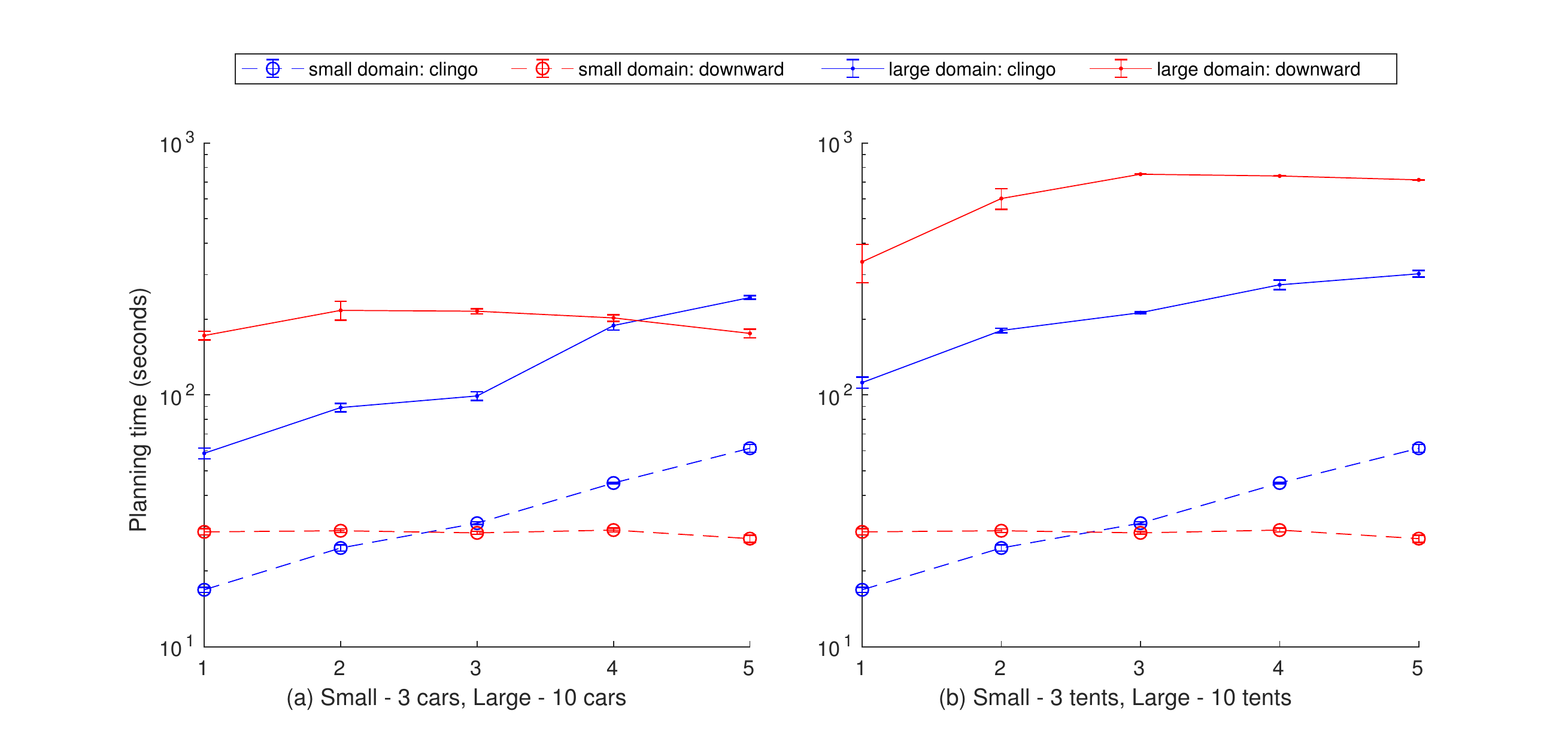}\\
    \includegraphics[width=.95\textwidth]{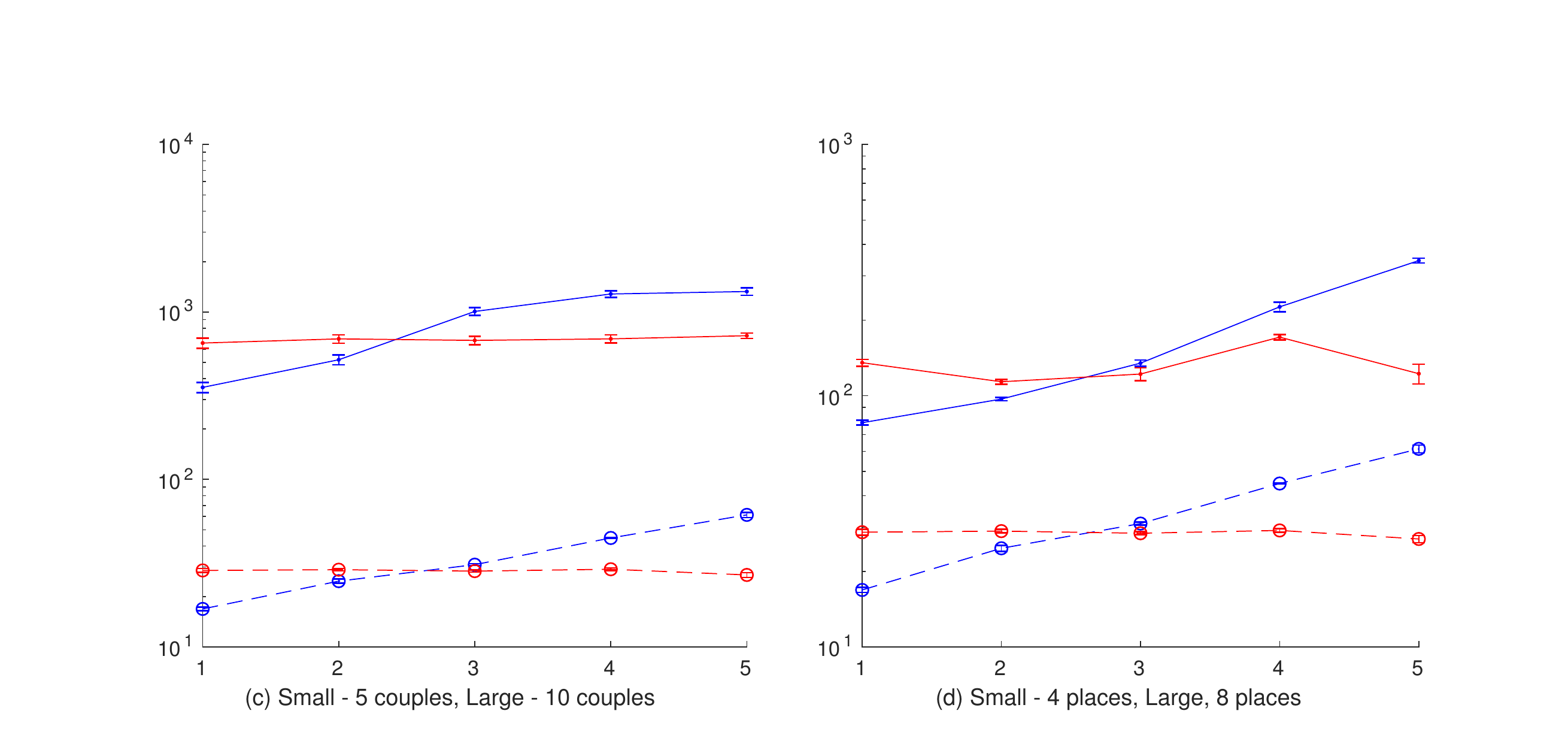} \vspace{-2pt}
    \caption{Varying individual domain attributes in the Hiking domain}
    \label{fig:more_hiking}
  \end{center}
  \vspace{-10pt}
\end{figure*}

\begin{figure*}[tb]
  \vspace{-2pt}
    \begin{center}\hspace{0em}
    \includegraphics[width=.7\textwidth]{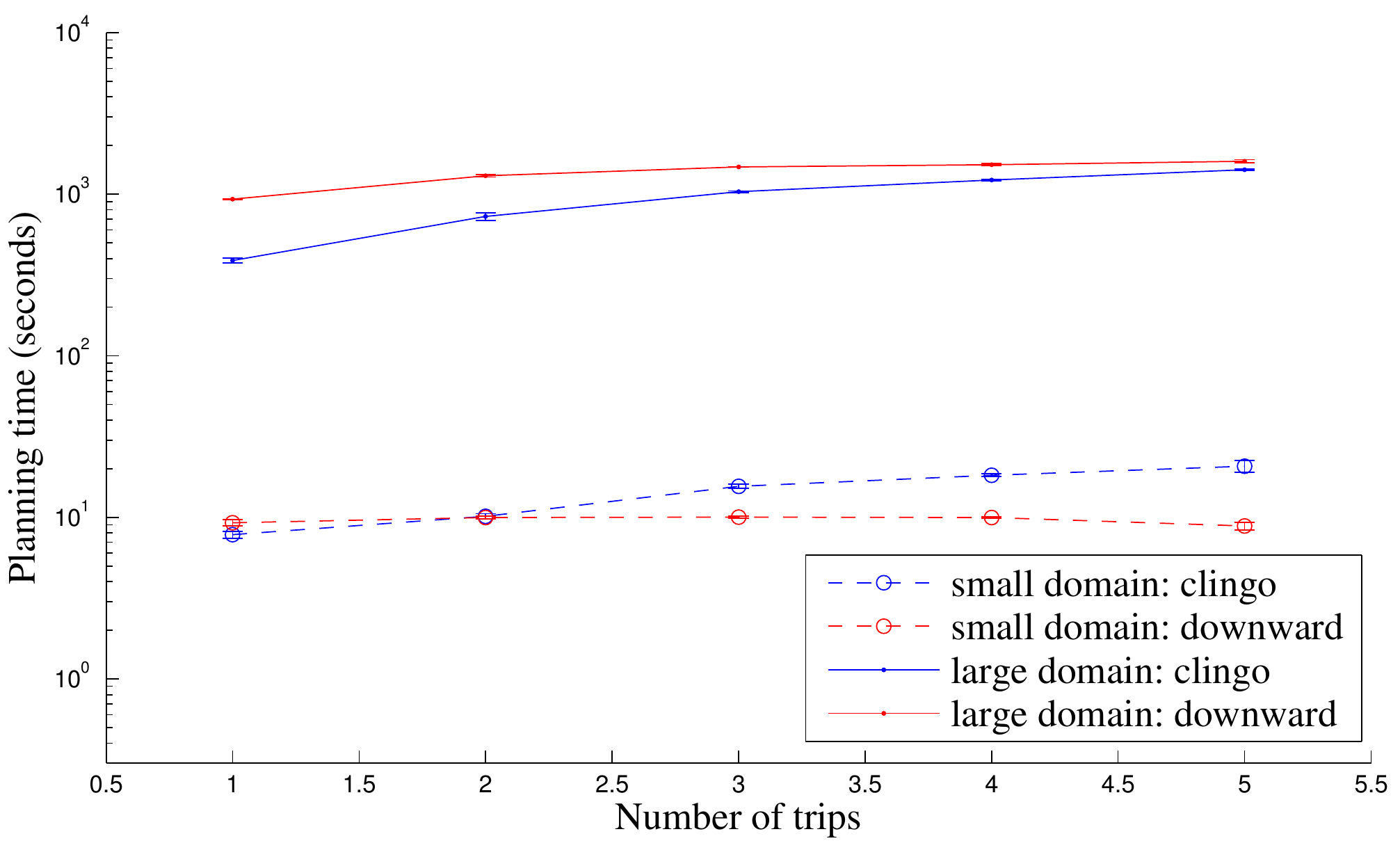} 
    \caption{Hiking: Small Domain vs. Large Domain}
    \label{fig:small_large}
    \end{center}
    \vspace{-10pt}
\end{figure*}

\paragraph{Blocks World} 
Figure~\ref{fig:blocksworld} reports results from the regular Blocks World domain.
Similarly, two versions of the domain are evaluated: a small domain with 15 blocks in the environment, and a large domain with 60 blocks in the environment.
Initially, all blocks are unstacked and on the table (so as to control the optimal plan length). 
The goal of the planners is to generate a plan that builds a single stack of a specified height with randomly specified blocks.
The plan length is proportional to the height of the stack, and is represented as the x-axis in Figure~\ref{fig:blocksworld}.
All results have been averaged across 10 trials.

The two red lines are far apart and almost horizontal, i.e.~ the \textsc{PDDL} planner slows down significantly in the larger domain, but planning time does not depend on plan length.
The planning time of Clingo grows as the plan length increases, but the difference between different sized domains is smaller than \textsc{PDDL}.
These observations support our hypotheses that \textsc{PDDL}-based planning is fast at producing long plans, while \textsc{ASP}-based planning is faster in large domains where smaller plans are necessary.

Figure~\ref{fig:blocks_extended} compares the performance of each planner between the regular Blocks World domain and the extended version.
In this experiment, the domain has 10 blocks. The task is the same as before: building a tower of various heights from unstacked blocks. The difference of the extended domain is that the goal condition is expressed with the predicate \c{above}, where the planners reason about recursive effects.
Similarly, the results are averaged across 10 trials.

In Figure~\ref{fig:blocks_extended}, there is a small difference in \textsc{ASP} solving time, but a significant change in the performance of the \textsc{PDDL} planner. In contrast to Figure~\ref{fig:blocksworld}, we can observe that complex reasoning affects \textsc{PDDL} planning time in its slope with respect to plan length, whereas object scaling shifts the curve up. The observation confirms that \textsc{ASP}-based planning is better for domains that involve complex reasoning (reasoning about recursive action effects in this case).



\paragraph{Hiking}
Figure~\ref{fig:more_hiking} and~\ref{fig:small_large} show results from the Hiking domain.
In each graph, the x-axis is the total number of trips made by all couples, and the y-axis is the planning time in seconds.
Figure~\ref{fig:small_large} shows the performance of each planner for two sizes of the domain.
Although we can make similar observations that the \textsc{PDDL} planner becomes faster than the \textsc{ASP} planner at higher plan lengths but slower with more objects, the evidence is weaker than the results above.
Since the Hiking domain has four types of objects that affect planning in different ways, the domain size has four dimensions.
Figure~\ref{fig:more_hiking} shows a more controlled set of experiments in which the larger domain increases only one type of objects.
Objects are added in a way that does not affect the plan length or the number of optimal plans at each point of x-axis.
In all graphs the slope of blue curves is larger than the slope of red curves.
So the hypothesis about the plan length holds for all object types.

When adding cars and tents to the domain, from Figure~\ref{fig:more_hiking}(a) and \ref{fig:more_hiking}(b), we can observe that the \textsc{ASP} solver is less sensitive than the \textsc{PDDL} planner, but there is no significant difference in the case of couples and places -- see Figure~\ref{fig:more_hiking}(c) and \ref{fig:more_hiking}(d).
We find that couples and places are more heavily used as action parameters than other two types of objects.
Actions such as \c{drive\_passenger} and \c{walk\_together} even take two parameters of each.
Therefore, increasing the number of these objects complicates the grounding\footnote{Plan generation in ASP is a two-step process that includes \emph{grounding} and \emph{solving}, where the grounding step outputs a variable-free representation.} of \textsc{ASP} problems.
We also observe that most of the increased planning time of Clingo is in grounding.
For instance, when the number of couple increases from 5 to 10 -- see Figure~\ref{fig:more_hiking}(c), average solving time for one trip stays below 0.8 second, while average grounding time grows from 16.0 seconds to 354.2 seconds.
Based on these two observations, we hypothesize that if the domain has actions that check or change the state of many objects, Clingo's advantage at planning in large domains can be canceled out by the extra grounding time.
A further analysis of how parameter type and number affect grounding time is left for future research.


\paragraph{Remark}

All of the above results support our hypothesis that \textsc{PDDL}-based approaches work better in situations where the generated plans have many steps, and \textsc{ASP}-based approaches work better in situations where the domain is large or substantial reasoning is required at every step of the plan to compute the world state. 

Given the wide range of hardware configurations of machines in the HTC cluster and the low standard deviation values reported in the results, we can see machines of different configurations did not cause significant differences in planning time, which indicates that the trends are consistent across different types of computing platforms.

\section{Conclusions and Discussions}
\label{sec:discussion}

In this article, we empirically compared \textsc{ASP}-based and \textsc{PDDL}-based task planners using three robotic benchmark domains. 
PDDL is the dominant action language in the task planning community; ASP is widely used for knowledge representation and reasoning, and can be used for task planning. 
The analysis in this article demonstrates that \textsc{PDDL}-based planners perform better when tasks require long solutions.
However, \textsc{ASP}-based task planners are less susceptible to an increase in the number of domain objects, as long as the growth in the number of objects does not explode the number of grounded actions.
Finally, in domains requiring complex reasoning such as the Robot Navigation domain, \textsc{ASP}-based planners can be considerably faster than \textsc{PDDL}-based planners for shorter plans.
Such observations can serve as a useful reference to task planning practitioners in the process of action language selection. 

This article, by no means, aims to provide a list of the best planners given a pair of planning domain and planning problem, which is infeasible in practice. 
From a high-level perspective, the observations shared in this article are intended to contribute to the community's understanding of how properties of planning domains and problems affect the performance of different planning systems. 
Based on one's knowledge and intuition about properties of the planning problems at hand, a practitioner can then make a more informed choice of the planning system.

In this work, we selected the Clingo system for evaluating the ASP-based planning formalism, and the FastDownward planning systems of FDSS-1 and LAMA-2011 for the PDDL-based formalism. These award-winning optimal systems represent the state-of-the-art algorithms and implementations of the two planning paradigms. 
As the first work on empirical comparisons of PDDL-based and ASP-based task planners, we focus on a clear presentation of the methodology and results.
We carefully selected the three domains that are as distinct as possible for a representative comparison. We believe the conclusions hold in most cases. 
In the future, we will conduct further evaluations over the two planning formalisms and their implementations on other task planning problems and using other more extreme computing platforms (such as low-end onboard computers on mobile robots). 
  The experiments in this article were conducted using an ASP-based and
  a PDDL-based planner that are meant to be representative of their
  respective classes.  While we believe that our results and
  observations will generalize to other such planners, we acknowledge
  that there is no way to establish that conclusively without
  empirically comparing with many other planners, which is beyond the
  scope of this article. 


This article does not include formal analysis or comparisons between PDDL-based and ASP-based paradigms. 
One of the main reasons is that the original definition of the PDDL language includes only its syntax, and does not discuss its semantics. As a result, different PDDL solvers have different ``interpretations'' of PDDL programs, although they all aim at producing optimal solutions. 
Another direction for future work is to look into the semantics of PDDL~\citep{mcdermott2003formal,thiebaux2005defense}, and conduct formal analysis between the two planning paradigms. 

\section*{Acknowledgements}
A portion of this work has taken place in the Learning Agents Research
Group (LARG) at UT Austin.  LARG research is supported in part by NSF
(IIS-1637736, IIS-1651089, IIS-1724157), ONR (N00014-18-2243), FLI
(RFP2-000), Intel, Raytheon, and Lockheed Martin.  Peter Stone serves on
the Board of Directors of Cogitai, Inc.  The terms of this arrangement
have been reviewed and approved by the University of Texas at Austin in
accordance with its policy on objectivity in research.

\bibliographystyle{fitee}
\bibliography{references}

\end{document}